\documentclass{article}
\usepackage{arxiv}

\usepackage{hyperref}       
\usepackage{url}            
\usepackage{booktabs}       
\usepackage{amsfonts}       
\usepackage{nicefrac}       
\usepackage{microtype}      
\usepackage{lipsum}		    
\usepackage{graphicx}
\usepackage[font={small,sf,bf}]{caption}
\usepackage{amsmath,amssymb,amsfonts}
\usepackage{xcolor}
\usepackage{diagbox}
\usepackage{colortbl}
\usepackage{multirow}
\usepackage{enumerate}
\usepackage{siunitx}
\usepackage{subfig}
\usepackage{float}
\usepackage{pict2e}

\newcolumntype{M}[1]{>{\centering\arraybackslash}m{#1}}
\newcolumntype{N}{@{}m{0pt}@{}}

\newif\iffinal

\iffinal
 \newcommand{\zliu}[1]{}
 \newcommand{\ian}[1]{}
\else
 \newcommand{\zliu}[1]{{\textcolor{blue}{ Zhengchun: #1 }}}
 \newcommand{\ian}[1]{{\textcolor{red}{ Ian: #1 }}}
\fi

\author{
Zhengchun Liu\\
Data Science and Learning division\\
Argonne National Laboratory\\
Lemont, IL 60439 \\
\texttt{zhengchun.liu@anl.gov} \\

\And
Ahsan Ali\\
Data Science and Learning division\\
Argonne National Laboratory\\
Lemont, IL 60439 \\

\And
Peter Kenesei \\
X-ray Science Division\\
Argonne National Laboratory\\
Lemont, IL 60439 \\

\And
Antonino Miceli \\
X-ray Science Division\\
Argonne National Laboratory\\
Lemont, IL 60439 \\

\And
Hemant Sharma \\
X-ray Science Division\\
Argonne National Laboratory\\
Lemont, IL 60439 \\

\And
Nicholas Schwarz \\
X-ray Science Division\\
Argonne National Laboratory\\
Lemont, IL 60439 \\

\And
Dennis Trujillo \\
X-ray Science Division\\
Argonne National Laboratory\\
Lemont, IL 60439 \\

\And
Hyunseung Yoo \\
Data Science and Learning division\\
Argonne National Laboratory\\
Lemont, IL 60439 \\

\And
Ryan Coffee \\
SLAC National Accelerator Laboratory\\
Menlo Park, CA 94025 \\

\And
Naoufal Layad\\
Stanford University\\ 
Stanford, CA 94305, USA\\

\And
Jana Thayer\\
SLAC National Accelerator Laboratory\\
Menlo Park, CA 94025 \\

\And
Ryan Herbst \\
SLAC National Accelerator Laboratory\\
Menlo Park, CA 94025 \\

\And
Chun Hong Yoon \\
SLAC National Accelerator Laboratory\\
Menlo Park, CA 94025 \\

\And
Ian Foster \\
Data Science and Learning division\\
Argonne National Laboratory\\
Lemont, IL 60439 \\
}

\date{}

\renewcommand{\shorttitle}{Bridging Data Center AI Systems with Edge Computing}

\title{Bridging Data Center AI Systems with Edge Computing for Actionable Information Retrieval}

\begin{document}

\maketitle

\begin{abstract}
Extremely high data rates at modern synchrotron and X-ray free-electron laser light source beamlines motivate the use of machine learning methods for data reduction, feature detection, and other purposes. 
Regardless of the application, the basic concept is the same: data collected in early stages of an experiment, data from past similar experiments, and/or data simulated for the upcoming experiment are used to train machine learning models that, in effect, learn specific characteristics of those data; these models are then used to process subsequent data more efficiently than would general-purpose models that lack knowledge of the specific dataset or data class.
Thus, a key challenge is to be able to train models with sufficient rapidity that they can be deployed and used within useful timescales.
We describe here how specialized data center AI (DCAI) systems can be used for this purpose through a geographically distributed workflow. 
Experiments show that although there are data movement cost and service overhead to use remote DCAI systems for DNN training, the turnaround time is still less than 1/30 of using a locally deploy-able GPU.
\end{abstract}

\section{Introduction}
The increased coherence and brilliance of next-generation X-ray light sources, such as the upgraded Advanced Photon Source and the Linac Coherent Light Source, APS-U and LCLS-II respectively, will lead to increasingly large and rich data sets produced at high data rates. 
Such multi-modal data, captured in situ, can provide new insights into rare events such as crack initiation and phase transformations. 
However, the complexity and velocity of these data means that extracting the desired physical information becomes a considerable computing challenge, particularly when information is needed rapidly, for example to steer experiments. 
This data extraction challenge is not easily addressed by using conventional analytical methods. Such methods take too long to run on individual processors, and even if efficiently parallelizable (factors like irregular memory accesses often make them latency bound), they still need many processors to complete analyses in near real-time~\cite{bicer2017trace}. 
Fortunately, machine learning (ML)-based surrogate models can provide an effective alternative to conventional methods in these contexts.
A suitably trained ML surrogate can approximate the results of an analytical method with high accuracy, and while it may perform more floating-point operations that the analytical method, can be executed at high speeds on specialized AI accelerators such as GPU, TPU, and NPU \cite{liu2020braggnn}, which are sufficiently inexpensive to deploy near to data sources.

There is a growing body of work on AI/ML methods for processing of X-ray source data for tomography~\cite{jimaging4110128,liu2020tomogan,liu2019deep}, serial crystallography~\cite{ke2018convolutional,souza2019deepfreak}, X-ray diffraction~\cite{oviedo2019fast,vecsei2019neural}, and source diagnostics themselves~\cite{AlvaroNatComm,Audrey2019}. 
For example, Pelt et al.~\cite{jimaging4110128} and Liu et al.~\cite{liu2020tomogan,liu2019deep} used deep convolutional neural networks to improve tomographic reconstruction from limited measurements (e.g., sparse projections, short exposure time, or limited angle).
Others have used DNNs to guide data collection under budgeted dosage~\cite{9409314} and to enhance streaming tomography~\cite{liu2020tomogan,liu2019deep}.
Using a pre-trained DNN model as a prior, Aslan et al. incorporated learned priors into the generic reconstruction framework for the joint ptycho-tomography problem~\cite{aslan2021joint}.
Abeykoon el al. have deployed models to AI accelerators (such as edge TPU and Jetson) for low-cost data processing at the edge~\cite{abeykoon2019scientific}. 
These methods, when combined with AI accelerators, thus make it feasible, in principle, to analyze data rapidly, at or near the point of data acquisition, rather than streaming them to a remote HPC system.

In order to use ML models effectively, we also need to be concerned with \textit{retraining}, for example when a sample or experiment setup changes.
\emph{Training} involves the use of one of a variety of techniques to learn, from supplied training data (often a collection of feature--target pairs), a mapping function from features to targets. 
The output of this process is a trained model that can then be applied to other features (a process often referred to as \emph{inference}) to obtain an estimation of the corresponding targets.
In the case of deep neural networks (DNNs), training uses a computationally intensive process called stochastic gradient descent to solve the optimization problem of finding good DNN parameter values.

\section{Motivations, Solution, and Gaps}
\begin{figure*}[h]
\centering
\includegraphics[width=.9\linewidth]{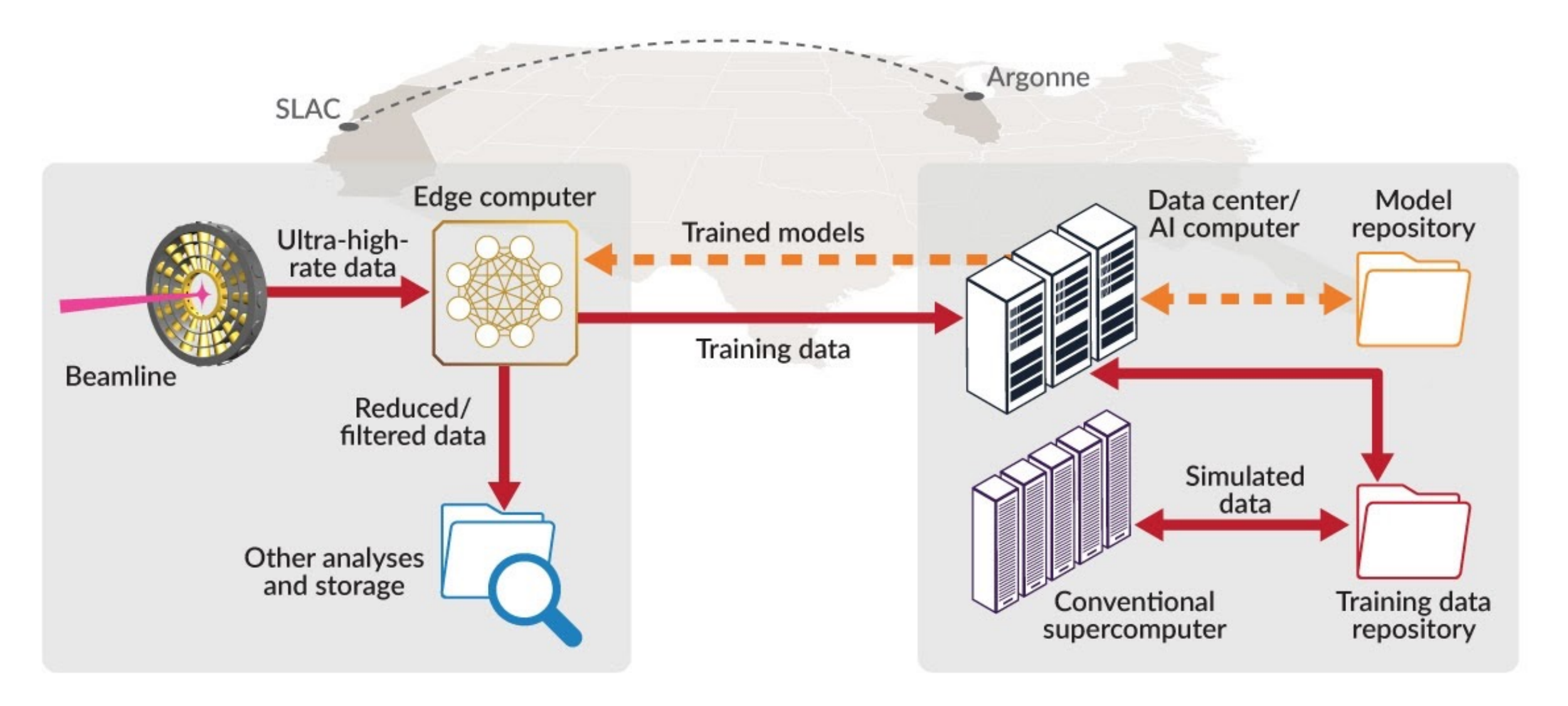}
\caption{A demonstration of the actionable information architecture uses an edge computer co-located with the experimental apparatus (CookieBox indicated here, the SLAC LCLS-II TMO beamline) for rapid reduction/filtering of high-velocity data; the ML model used to perform reduction/filtering is trained on a remote data center AI computer (here, at the Argonne Leadership Computing Facility). Solid crimson arrows are data flows; dashed orange are models.}
\label{fig:bigpic}
\end{figure*}

We define a data center AI (DCAI) system as an AI accelerator that must be deployed in a data center due to its cooling, power supply, ventilation, and fire suppression requirements.
DCAI systems can train ML models much more rapidly than computing clusters that are maximally deploy-able within the experiment facility.
Once the DNN is trained, we use another set of AI accelerators specialized for model inference, called \texttt{edge-AI} (e.g., FPGA, GPU with Tensor cores, edge TPU \cite{abeykoon2019scientific}), to process experiment data near the data acquisition in real-time. 
Since model inference is much less computing intensive than model training and it only needs to be as fast as the data generation rate (i.e., real/wall- time speed), \texttt{edge-AI} can be lightweight enough to be deployed within the experiment facility.
Thus, in order to bridge the divide of powerful AI systems deployed in data center (e.g., scientific computing facility) and rapid ML model (re)training needs at experimental facility, we need to have a workflow, as shown in \autoref{fig:bigpic}, that is geographically distributed on multiple science facilities to (re)train ML models rapidly.

The basic concept of utilizing ML for real-time actionable information retrieval is the same irrespective of the application, as illustrated in \autoref{fig:bigpic}: data generated by a simulation close to the experiment, and/or collected in early stages of an experiment, and/or data from past similar experiments, are used to train models that, in effect, learn specific characteristics of the supplied training data; these models can then be used to process subsequent data more efficiently than general-purpose models that lack knowledge of the specific datasets or classes of data \cite{liu2019deep, liu2020braggnn, cherukara2020ai}.
For example, in High-Energy X-Ray Diffraction Microscopy (HEDM), a single scan of 1440--3600 frames can be obtained in 5--10 minutes now and in perhaps 50--100 seconds at APS-U. 
When measuring a single sample on a layer-by-layer basis, similar data quality is observed repeatedly. 
Thus, an AI model trained on early layers can be used to process latter layers.
In serial crystallography~\cite{Nass2020,Meents2017}, Bragg spots from past experiments can be used for training detectors for use in the current experiment. 
In single particle imaging \cite{Seibert2011,Hosseinizadeh2017}, the particles are unique for each experiment; realistic simulations can be used to train a neural network for classifying images of interest. 
In ptychography, the diffraction patterns collected at early stages of an experiment and their corresponding phase retrieved using the conventional solution of an inverse problem can be used to train a ML model, such as PtychoNN \cite{cherukara2020ai}, and the trained model can then be used for more rapid phase retrieval from subsequent diffraction patterns.

Such applications in which a DNN is (re)trained for actionable information retrieval at the edge, near to a data sources, usually has time constraints that require us to deliver the model within a limited time-frame, particularly in the case of in-situ experiments.
A key challenge is thus to be able to train models quickly enough that they can be deployed and used within useful timescales. 
The emergence of purpose-built accelerators (e.g., TPU~\cite{google-tpu}, Cerebras\cite{cerebras}, SambaNova\cite{sambanova}) for high-speed, high-velocity, and/or big data AI models allows for scenarios in which model training is performed on a specialized accelerator while inference is performed at the edge, near the data source, for real-time data analysis. 
However, usually it is not feasible to deploy these AI systems near an experiment facility because AI systems usually require a significant amount of hardware and software infrastructure, including power subsystems, stable and uninterruptible power supplies, proper ventilation, high-quality cooling systems, fire suppression, reliable backup generators, and connections to external networks.
Instead these AI systems are usually deployed in a data center, for which reason we refer to them as data center AI (DCAI) systems.
Moreover, there is also a strong economical argument of using DCAI systems, i.e. allowing to share the very expensive specialized AI processors between experiments in multiple facilities. 

In this paper, we propose an approach to the use of DCAI systems can be integrated into geographically distributed \textbf{workflow} solutions to facilitate model training on remote DCAI systems and model deployment on edge devices. 
Thus, one of the important elements to retrieve actionable information in real time is the use of powerful DCAI systems to enable rapid (re)training of ML/AI models that are then deployed on edge devices for production use, as shown in \autoref{fig:bigpic}.
A second major driver for the use of data center computers is for the generation of simulated data for model training.
Here we face two distinct needs: high-throughput runs to generate large quantities of simulation data, and low-latency runs to generate results rapidly during an experiment.

\section{Workflow Design and Implementation}
We used the Globus Flows~\cite{chard2019globus}, funcX~\cite{chard2020funcx}, and Globus file transfer~\cite{foster2011globus} services to build a workflow to automatically (re)train DNNs with given data or simulations using DCAI and HPC.
In this section, we briefly introduce key features and functionalities of each service and describe how the geographically distributed workflow is realized using these~services. 

The Globus \textbf{Flows} service introduces \textit{Action Providers}, \textit{Actions}, and \textit{Flows} to create custom processes solving particular research data management problems~\cite{chard2019globus,globus-service}. 
An \textit{Action Provider} is an HTTP accessible service which acts as a single step in a process and implements the Action Provider Interface.
An \textit{Action} represents a single, discrete invocation of an Action Provider. 
A \textit{Flow} represents a single process that orchestrates a series of services/actions into a self contained operation. 
One can think of a Flow as a declaratively defined ordering of Action Providers with condition handling to define expected success or failure scenarios.
Globus Auth~\cite{tuecke2016globus} is used to authenticate all interactions with Action Providers, Actions and Flows.

\textbf{funcX} provides the function-as-a-service capability to execute functions across a federated ecosystem of funcX endpoints~\cite{chard2020funcx}.
It basically offers the ability to turn any computing resource, including clouds, clusters, supercomputers, edgeAI devices and DCAI systems into a function-serving endpoint by deploying a Python-based funcX endpoint to the target system.
So we can then build our computation actions, including simulation, data annotation and model training, using funcX service and wrap it as an Action of Globus Flows. 
The Flow engine will then interact with funcX service to automatically execute the function (action).
More importantly, funcX is server-less and provides the fire-and-forget convenience for user. 

The \textbf{Globus} transfer service provides a secure, unified interface to the data. 
It allows us to start and manage transfers between endpoints, while automatically tuning parameters to maximize bandwidth usage, managing security configurations, providing fault recovery, and notifying users of completion and problems.
We use Globus to transfer data and trained DNNs between systems within and across organizations by wrapping each file transfer request as an action of Globu~Flows.

\begin{figure*}[htb]
\centering
\includegraphics[width=0.8\linewidth]{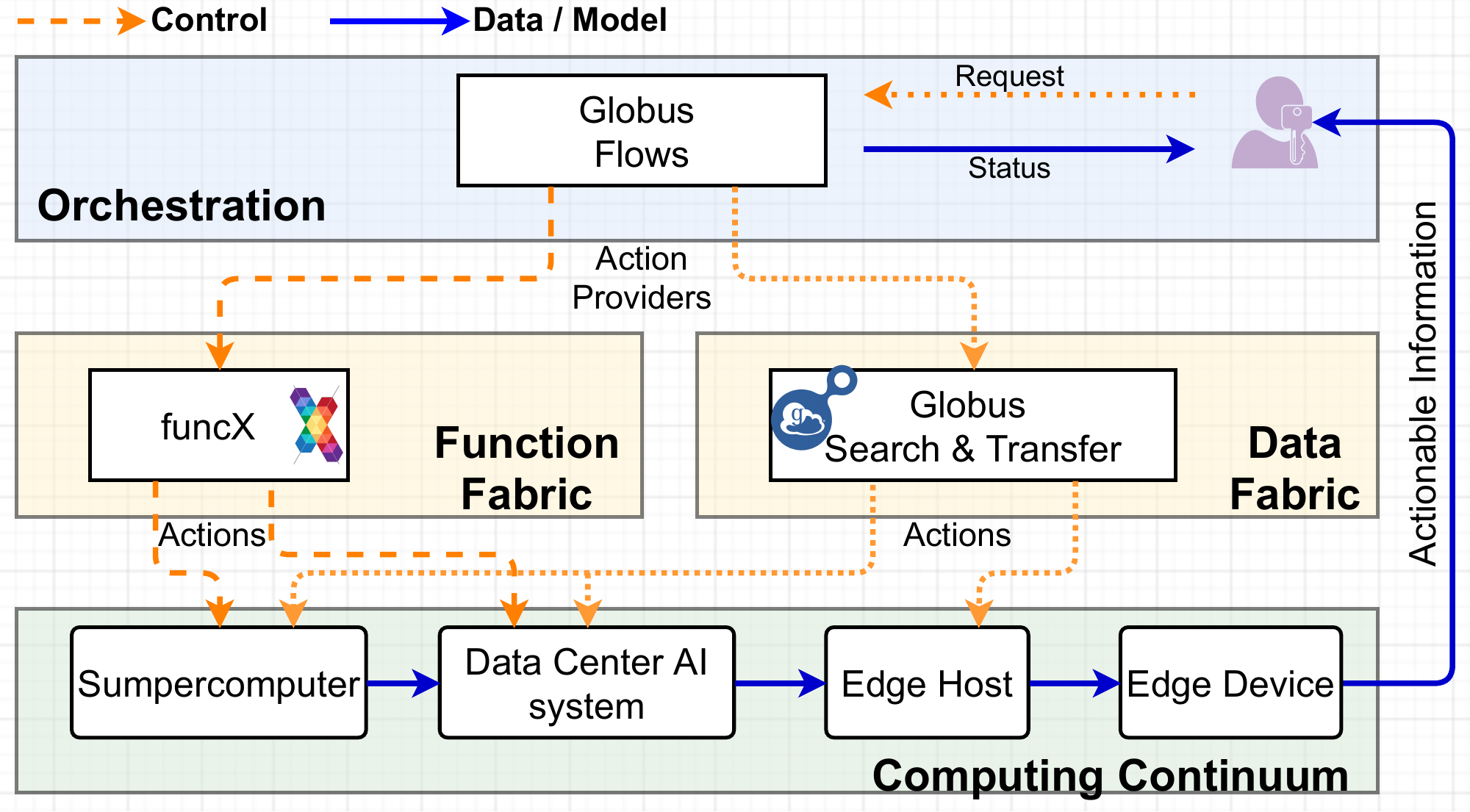}
\caption{Architecture and building blocks of the workflow. Solid arrows are data flows; dashed arrows are control flows.}
\label{fig:arch}
\vspace{-2ex}
\end{figure*}

\autoref{fig:arch} shows the overall architecture of the system.
Basically, all computing functions, e.g., simulation for training data generation, data curation and model training, are abstracted as a funcX function and wrapped as an action of Globus Flows; and all data dependencies are solved by wrapping the data transfer as an action of Globus Flows using Globus file transfer service.
A developer builds and deploys the workflows to the \textbf{Flows} service and shares it with actual users (e.g., experimental scientists).
The user only needs to interact with the workflows through a client to initiate the training task and the flow engines will orchestrate resources with action providers to run the workflow and deliver the trained model to the place that is defined in the workflow by user.

The implementation of the workflow shown in this paper is available at \url{https://github.com/AISDC/DNNTrainerFlow} and one can use it as a reference to rebuild their own workflow if similar building blocks are used.
A recording of the demonstration of the workflow that bridges SLAC experimental facility and Argonne Leadership Computing Facility(ALCF) is available at \url{https://youtu.be/lL6HsIk3xjE}.

\section{Analytical Modeling and Evaluation}
\subsection{Performance Modeling}\label{sec:model}

We construct the various applications that we consider here in terms of the following six basic operations:
\begin{enumerate}[i)]
\item \textbf{C}ollect a datum, $d$ or  
\item \textbf{S}imulate an experiment to generate a datum, $d$, without an experiment;
\item \textbf{A}nalyze a datum using a conventional algorithm (e.g., Bragg peak localizing using pseudo-Voigt), generating an analysis (e.g., Bragg peak locations), $a$; 
\item \textbf{T}rain (or retrain) a ML model with some number of \{$d$, $a$\} pairs, generating a model, $m$;
\item \textbf{D}eploy the ML model, $m$, to an edge-AI device; and
\item \textbf{E}stimate an analysis with a previously trained ML model, generating an estimated analysis, $\hat{a}$.
\end{enumerate}

When an operation may be performed at different locations, we denote that by a subscript, such as \textbf{A}\textsubscript{ex} for an analysis performed on a computer at an experiment and \textbf{A}\textsubscript{dc} for an analysis performed in a data center.
We represent the movement of data $d$ from location $a$ to location $b$ as $a\xrightarrow{d}b$. 
Finally, we define $\mathcal{C}(o)$ to be the cost of an operation $o$.
Thus, for example, the conventional way of moving data from experiment to data center for analysis, and then returning result to the experiment (e.g., for steering), will cost, as a function of $d$
\begin{equation}
f^{c}(d) = \mathcal{C}(ex\xrightarrow{d}dc) + \mathcal{C}\left(\textbf{A}\textsubscript{dc}(d)\right) + \mathcal{C}(dc\xrightarrow{a}ex),
\label{eq:streaming}
\end{equation}
while performing analysis at the experiment facility will cost 
\begin{equation}
f(d) = \mathcal{C}(\textbf{A}\textsubscript{ex}), 
\label{eq:all-local}
\end{equation}
and estimating an analysis at the experiment will cost $\mathcal{C}$(\textbf{E}\textsubscript{ex}).

In comparison for processing the same amount of data $d$, we first move a data subset, $\bar{d}$, from experiment to data center for analysis and train a model, and then returning a model for \textbf{E}stimating the subsequent analysis will cost, as a function of $d$:  
\begin{equation}
\begin{aligned}
f^{ml}(d) = \mathcal{C}(ex\xrightarrow{\bar{d}}dc)) + \mathcal{C}(\textbf{A}\textsubscript{dc}(\bar{d})) + \mathcal{C}(\textbf{T}\textsubscript{da}(\bar{d})) + \\
\mathcal{C}(dc\xrightarrow{m}ex) + \mathcal{C}(\textbf{E}_{d - \bar{d} })
\end{aligned}
\label{eq:aisdc}
\end{equation}

The costs for \textbf{C}, \textbf{S}, \textbf{A}, and \textbf{E} are deterministic when dedicated resources are used and thus only need to be profiled once for a given type of experiment.
The cost of \textbf{T} depends on the specific AI system used, it can range from seconds to hours, as we will discuss in \S\ref{sec:exp}.
Wide-area network congestion~\cite{hpdc17-zliu, hpdc18-zliu} can result in variability in 
the data movement cost, $\mathcal{C}(a\xrightarrow{d}b)$, but 
most research and education networks (e.g., ESnet and Internet2) are over-provisioned in order to support capacity bursts~\cite{huang2005does,Internet2,esnet}. 
For example, when the backbone circuit of Internet2 is seen regularly to sustain 40\% utilization, staff initiate a backbone augment~\cite{Internet2}.
Moreover, previous work shows that the model for $a\xrightarrow{t}b$ can be approximated by using a linear model $T = \frac{1}{v}x + S$ where $x$ is the bytes in total, $v$ is average transfer rate and $S$ is a constant start up cost mainly depends on the number of files in the dataset~\cite{liu2021design,liu2019data}.
More complicated machine learning based prediction model can also be incorporated for more congested network and data/or data transfer nodes~\cite{liu2018building}.

\subsection{Model based Analysis}\label{sec:model-ana}
The analytical model proposed in \S\ref{sec:model} can be used to evaluate different data analysis solutions (e.g., Equations~\ref{eq:streaming}--
\ref{eq:aisdc}) for a given experiment.
As the cost of \textbf{T}, we will demonstrate the evaluation by using HEDM as an example in \S\ref{sec:exp}.

Here we use BraggNN~\cite{liu2020braggnn}, an ML-based surrogate solution for Bragg peak analysis (see \S\ref{sec:dnns}), as an example to evaluate the most proper way of data analysis. 
As reported by Liu et al.~\cite{liu2020braggnn}, \textbf{A} takes about 2000 core$\times$seconds for \num{800000} peaks, while \textbf{E} with BraggNN can process \num{800000} peaks in 280 ms (batch processing).
Here we assume that there is a CPU cluster with \num{1024} usable cores (to our knowledge, GPU accelerated \textbf{A} does not exist, mostly because its computing and memory access pattern do not fit well with GPU architecture).
Thus, we get $\mathcal{C}\left(\textbf{A}\textsubscript{dc}(d)\right)=2.44 \mu s$, and $\mathcal{C}\left(\textbf{E}\textsubscript{dc}(d)\right)=0.35 \mu s$ where $d$ denotes one patch of Bragg peak with 11 by 11 pixels, 16 bits per pixel.

\begin{figure}[htb]
  \setlength{\unitlength}{0.1\columnwidth}
  \begin{picture}(4,2.65)
  \put(0,0){\includegraphics[height=28mm,trim=0.3mm 0 0 0,clip]{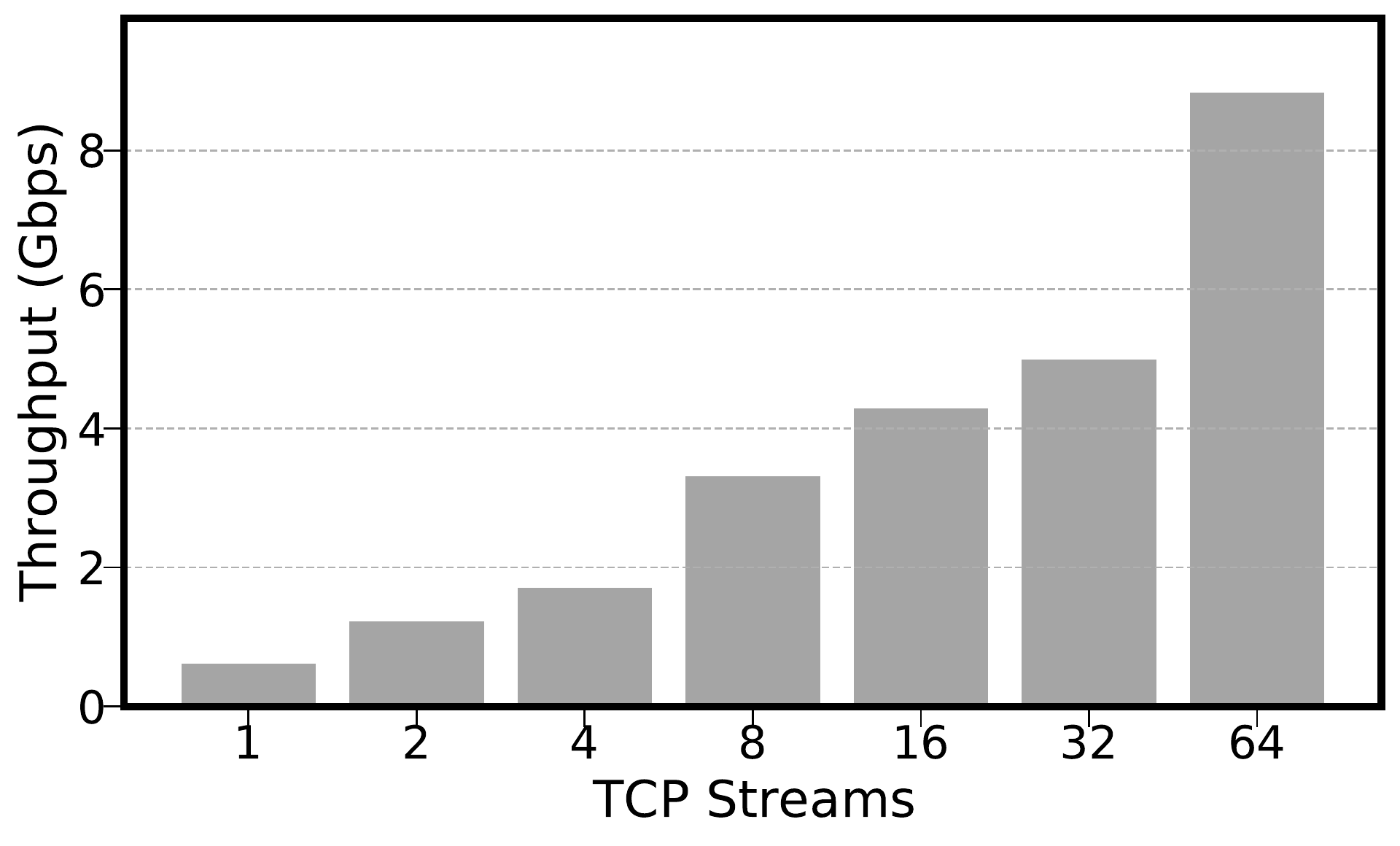}}
  \put(5.2,0){\includegraphics[height=28mm,trim=15mm 0 0 0,clip]{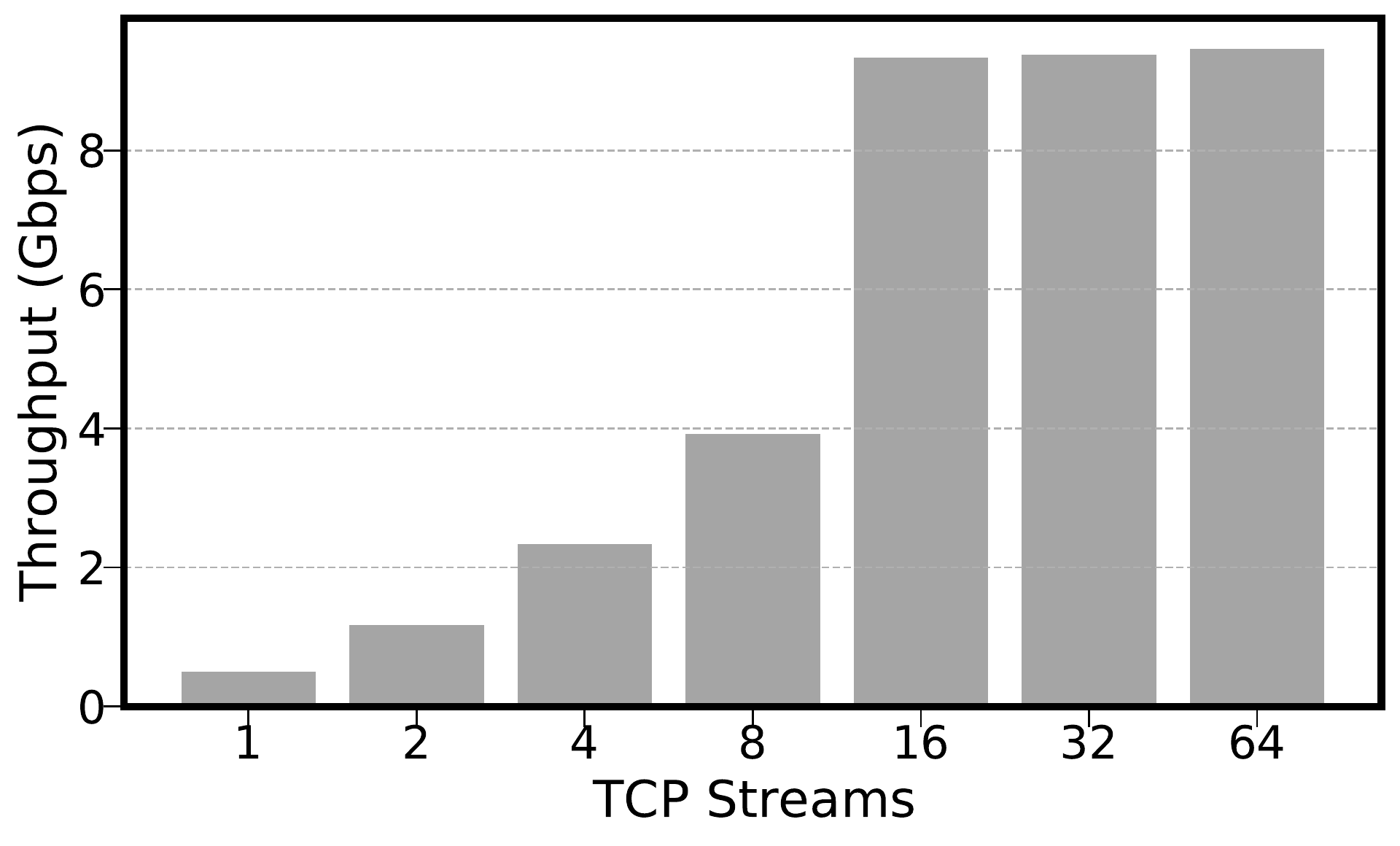}}
    \put(.6,2.7){\mbox{ALCF$\rightarrow$SLAC}}
    \put(5.4,2.7){\mbox{SLAC$\rightarrow$ALCF}}
  \end{picture}
\caption{File Transfer performance between ALCF and SLAC.}
\label{fig:trs}
\vspace{-.2cm}
\end{figure}

\begin{figure}[htb]
\centering
\includegraphics[width=0.4\textwidth]{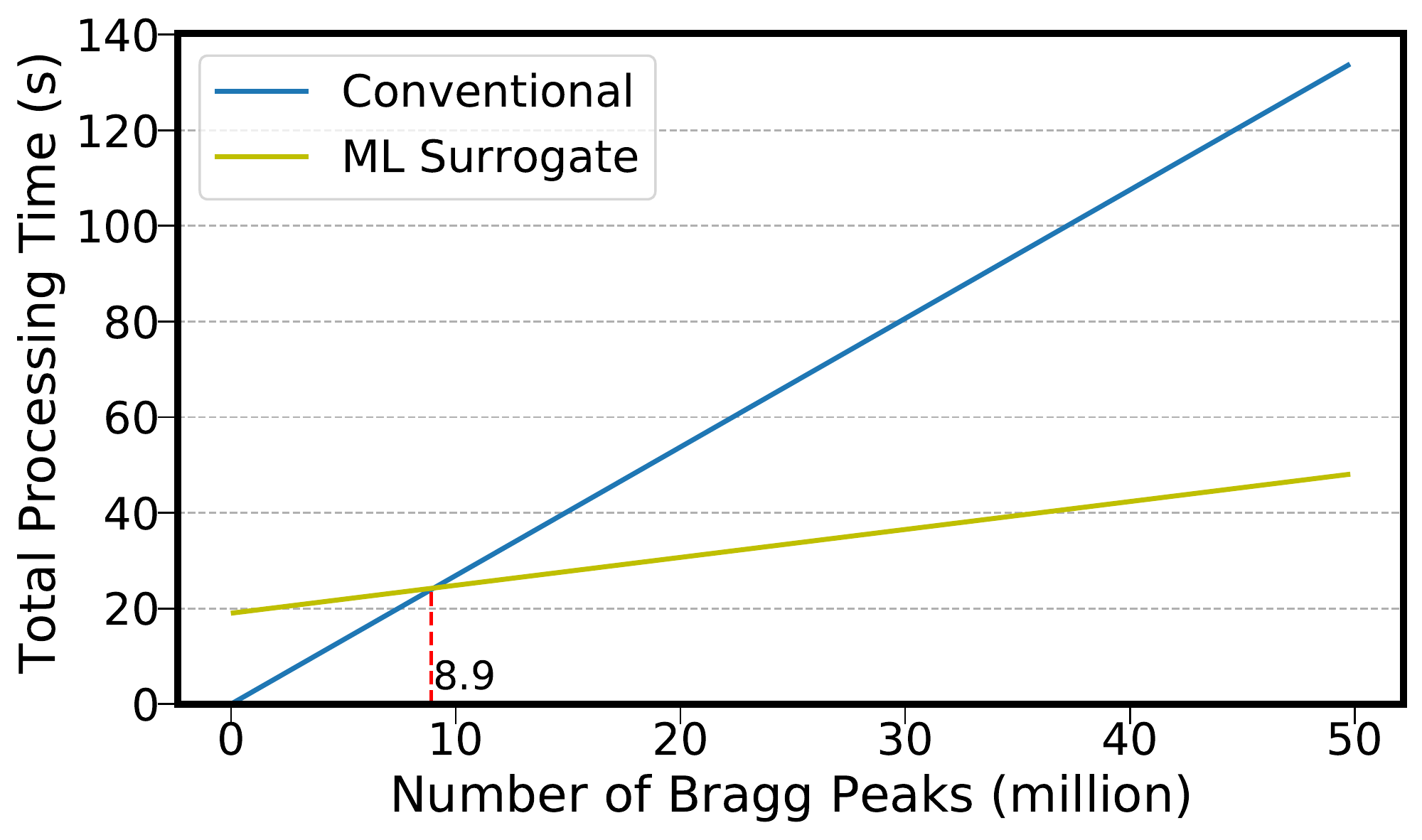}
\caption{A comparison of conventional  vs.\ ML-based surrogate solution for processing datasets of varying size.}
\label{fig:cmp}
\end{figure}

The backbone fiber connects experimental facility (e.g., SLAC here) and data center (e.g., ALCF here) is 100 Gbps.
We benchmarked the file transfer throughput using Globus with one DTN with a 10 Gbps network card at each side and presented results in \autoref{fig:trs} with different level of parallelism.
As one can see, we can get more than 1GB/s when transfer multiple files concurrently (detailed study in \cite{kettimuthu2018transferring}).
As discussed in \S\ref{sec:model}, the network is over-provisioned, we conservatively assume that we can get 1 GB/s at most time. 
Thus we have $\mathcal{C}(ex\xrightarrow{d}dc)) = \mathcal{C}(dc\xrightarrow{d}ex)) = 0.24 \mu s$.
As modeled in \autoref{eq:aisdc}, we need to transfer a portion, $p$, of our raw data (say $p$=10\%) to data center, label it using \textbf{A} and then train BraggNN with \textbf{T}.
Both the trained BraggNN model (3~MB, i.e., 3000~$\mu s$) and the labeled portion (each datum leads to 8 bytes) need to be transferred back to SLAC.
As will be discussed in \autoref{tbl:time-breakdown}, BraggNN can be trained in as little as 19 seconds when using the Cerebras system.
We thus assume $\mathcal{C}(\textbf{T}\textsubscript{da}(\bar{d}))=19s$.
If the dataset to be processed has $N$ datum, then the total time using conventional way with a cluster at data center will take
\begin{equation}
\begin{aligned}
f^{c}(d) = N * 0.24 + N * 2.44 + N * 8 * 10^{-3} \quad (\mu s),
\end{aligned}
\label{eq:streaming-obj}
\end{equation}
while the time when use BraggNN will be 
\begin{equation}
\begin{aligned}
f^{ml}(d) = {} & {} p * N * 0.24 + p * N * 2.44 \\& + p * N * 8 * 10^{-3} + 19*10^6 + 3000 \\& + (1-p) * N * 0.35 \quad (\mu s).
\end{aligned}
\label{eq:aisdc-obj}
\end{equation}
The number of Bragg peaks per experiment ranges from tens of hundred thousands to millions. We vary $N$ and compare conventional solution with ML surrogate based solution in \autoref{fig:cmp}.
As there is a static cost to train the BraggNN, the conventional solution outperforms proposed solution only when the number of data is small. 
The analytical model can thus be used to decide which solution to take before processing.

\section{Experiment and Discussion}\label{sec:exp}
In this section, we will demonstrate the workflow designed for rapid (re)training of two DNNs with the above discussed steps.
Here we care more about the end-to-end to (re)train a machine learning model with new dataset from a user's point of view.
The end-to-end is defined as the time difference from a user initializing the model (re)training with new dataset till the trained model is received at edge host machine of the user's choice.
We will compare the end-to-end time of the proposed solution using remote DCAI systems against an high-end GPU that can be deployed locally within the experiment facility.

\subsection{Experiment Setup}
This experiment involves a case in which SLAC needs to (re)train DNNs for an experiment.
We compare and contrast two scenarios: one in which model training is performed remotely, using a Cerebras or SambaNova DCAI system at ALCF, and one in which training is performed locally, on an NVIDIA V100 GPU co-located with the experiment. 

In the remote case, the DCAI system is 3000 km distant, with a network round trip time over the Energy Sciences Network (ESNet, a high-speed computer network serving United States Department of Energy scientists and their collaborators worldwide~\cite{esnet}) of about 48 ms at a peak bandwidth of 100 Gbps.
We must transfer the training dataset produced at SLAC to ALCF, train the model with the dataset, and transfer the trained model back to SLAC.
In the local case, there is no cost for wide-area data transfer. 


\subsection{Deep Neural Networks}\label{sec:dnns}
In order to demonstrate the concept of using remote DCAI for rapid model (re)training, we used two DNNs, BraggNN~\cite{liu2020braggnn} and CookieNetAE, as examples to demonstrate the superiority of using remote DCAI system for retraining. 
Both models, detailed as follows, are purposely designed to run inference at edge to retrieve actionable information from experimental data in real time, and both need retraining time to time.

\paragraph{CookieNetAE}
The CookieBox detector~\cite{Audrey2019} is an angular array of sixteen electron Time-of-Flight (eToF) spectrometers. 
The x-ray shot photo-ionizes gas molecules in the interaction point, ejecting electrons. 
These electrons drift through a series of electrostatic potential plates in the 16 channels and then detected by microchannel plates (MCP). 
This problem becomes difficult when we consider using a circularly polarized optical laser field in the interaction region and when the number of detected electrons is low.  
CookieNetAE is a deep neural network designed to estimates the energy-angle dependent probability density function of electrons' energy for all 16 channels. 
CookieNetAE takes an input image where each row in the image corresponds to an empirical energy histogram with 128 bins of 1~ev width of a given CookieBox channel built after the time-energy mapping. 
And the output will be an image containing the probability density of electrons’ energy in each channel.
 This model has 8 convolution layers totals 343,937 trainable parameters, rectifier activation function was used for all layers.
The mean squared error was used as the loss function and the Adam optimizer with a learning rate of 0.001 was used to train the model.
 
\paragraph{BraggNN}
 X-ray diffraction based microscopy techniques such as HEDM rely on knowledge of the position of diffraction peaks with high precision. 
These positions are typically computed by fitting the observed intensities in area detector data to a theoretical peak shape such as pseudo-Voigt. 
As experiments become more complex and detector technologies evolve, the computational cost of such peak detection and shape fitting becomes the biggest hurdle to the rapid analysis required for real-time feedback during in-situ experiments. 
BraggNN~\cite{liu2020braggnn} is a deep learning model designed to localize Bragg peak positions much more rapidly than conventional pseudo-Voigt peak fitting. 
Recent advances in deep learning method implementations and special-purpose model inference accelerators allow BraggNN to deliver enormous performance improvements relative to the conventional method, running, for example, more than 200 times faster than a conventional method on a consumer-class GPU card with out-of-the-box software.

\subsection{Results and discussion}\label{sec:res-discussion}
We conducted our experiments on three DCAI systems: multi-GPU server, Cerebras~\cite{cerebras} and SambaNova~\cite{sambanova} all located at ALCF.
Horovod~\cite{sergeev2018horovod} was used to implement DNN training using multi-GPU.
The entire wafer of the Cerebras was used for data parallelism via model replica.
\begin{table*}[htb]
\centering
\caption{Time breakdown of the workflow steps/actions when using either a remote Cerebras DCAI system, a remote 8-GPU server or a remote SambaNova (only used 1 out of 8 RDUs per node) versus using one local GPU.  The purpose of listing the performance of different systems is not to compare them, as this is not a systematical benchmarking study. The purpose is rather to demonstrate the feasibility of using powerful yet remote DCAI systems for AI at edge applications.}
\renewcommand{\arraystretch}{2}
\begin{tabular}{M{2.8cm}|M{1.8cm}|M{1.8cm}|M{1.8cm}|M{1.8cm}|M{1.8cm}}
\noalign{\hrule height 2pt}
{\bf Mode} \texttt{\textbackslash} {\bf Time} & \textbf{Neural Network} & \textbf{Data Transfer (s)} & \textbf{Model Training (s)} & \textbf{Model Transfer (s)} & \textbf{End-to-End (s)}  \\\noalign{\hrule height 1pt}
\textbf{Local (one GPU)}  & \multirow{3}{*}{BraggNN~\cite{liu2020braggnn}} & N/A            & 1102           & N/A              & 1102 \\\cline{1-1} \cline{3-6} 
\textbf{Remote (Cerebras,  Entire Wafer)}  &   & 7           & 19           & 5            & 31       \\\cline{1-1}\cline{3-6} 
 \textbf{Remote(SambaNova, 1-RDU)}  &   & 7           & 139           & 5            & 151        \\\noalign{\hrule height 1pt}
\textbf{Local (one GPU)}  & \multirow{3}{*}{CookieNetAE} & N/A           & 517           & N/A             & 517       \\\cline{1-1} \cline{3-6} 
\textbf{Remote (Cerebras, Entire Wafer)}  &   & 5           & 6           & 4            & 15       \\\cline{1-1}\cline{3-6}
\textbf{Remote (multi-GPU server)}  &   & 5          & 88           & 4            & 97        \\
\noalign{\hrule height 2pt}
\end{tabular}
\label{tbl:time-breakdown}
\end{table*}

\autoref{tbl:time-breakdown} presents the time breakdown of $\mathcal{C}(\textbf{T})$ for each of the actions/steps in the workflow and compares it with training the same model with the same dataset using a single local NVIDIA V100 GPU that can be deployed locally within the experiment facility.
As one can see, although the data and model transfer (i.e., cost when use remote resources) accounted nearly half of the end-to-end time in certain cases, using a remote DCAI system is still more than 30 times faster than using local resources for both BraggNN and CookieNetAE.

We note that only one (out of eight per node) Reconfigurable Dataflow Unit (RDU) is used for the BraggNN based experiment on SambaNova system.
Experiment for data parallelism over multiple RDUs is forthcoming. 
We did not conduct experiments for BraggNN using multi-GPU because BraggNN is light-weight by design that will be latency-bounded when perform data parallelism over multiple(distributed) GPUs, i.e., the speedup of computing gaining from using multiple GPUS is less than the necessary cost on gradients synchronization (\texttt{allreduce} operation).
We leave further optimization of BraggNN trained on multiple distributed GPUs for future work.


\section{Related work}
Real-time scientific experiment data analysis leveraging data center resource such as cloud or supercomputer has been an active area of research to fulfill the increasingly data-intensive analysis demanding from upgrades to science infrastructure. 
Wilamowski et al. \cite{wilamowski20212} used Globus Flows to automate real-time SSX data analysis, but did not use ML models or AI accelerators as here.
Rocki et al.~\cite{cs1forstencil} discussed the possibility of using DCAI systems  for PDE codes in scientific applications, and demonstrated the benefit over using conventional CPU or GPU based solutions.
Emani et al.~\cite{emani2021accelerating} explored the suitability of SambaNova, another DCAI system, for diverse AI for Science workloads and observed significant performance gains over traditional hardware. 
Acciarri et al.~\cite{acciarri2020cosmic} advanced state-of-the-art accuracy for an important neutrino physics image segmentation problem leveraging the large memory of DCAI systems which are not possible to fit on the highest-end GPU because of the large tensor size (convolutions neural networks with images beyond 50k x 50k resolution).

Thorpe et al.~\cite{thorpe2021dorylus} also used function-as-a-service computing (AWS Lambda) offered by cloud computing provider to scale and accelerate machine learning modeling training task.
In the context of end-to-end distributed science workflows, Salim et al.~\cite{salim2021toward} proposed the Balsam workflow framework to enable wide-area, multi-tenant, distributed execution of analysis jobs on DOE supercomputers.

\section{Conclusion and Future work}
We have presented an automated workflow for rapid deep neural networks training using remote resources, which automates the model (re)training and achieves a turnaround time between initiating and model delivery to edge host of less than 151 seconds including all data movement overhead.
As a comparison, it takes $\sim$17 minutes (though no cost on data movement) when training the same model using one high-end GPU that is deploy-able within the data acquisition machine.
This workflow proves the feasibility of using powerful yet remote data center AI systems to enable rapid (re)training of deep neural networks for production use on edge devices. 
We automated the rapid DNN training, using remote data center AI (DCAI) systems (e.g., Cerebras and SambaNova). 
The workflow also simplifies the use of remotely hosted DCAI systems for domain scientists. 

As for future work, there are three directions that we are currently pursuing. 
1), we are building the model repository, as shown in \autoref{fig:bigpic}, so as to pick up the right model as foundation to fine-tune using new dataset instead of retraining from scratch, to further accelerate the training process. 
2), we are also building a data repository to augment training dataset or substitute unlabelled dataset, because the labelling process, $ \mathcal{C}(\textbf{A}\textsubscript{dc}(\bar{d}))$, is usually time consuming (which motivates ML-based surrogate solution).
Both 1) and 2) will need data center (HPC, DCAI and cloud) resource.
3), in most cases the new experimental data to (re)train the DNN are not labelled, we will need to run labelling algorithm \textbf{A} (e.g., pseudo-Voigt profiling for BraggNN~\cite{liu2020braggnn}). 
As the training process is mini-batch based which can be started before getting all training samples, we can try to partially overlap \textbf{A} and \textbf{T} in the workflow to shorten end-to-end time.

\clearpage
\setcounter{section}{1}
\section*{\textbf{Appendix: Reproducibility}}
\subsection{Dependencies}
The workflow runs in the Globus\footnote{https://www.globus.org} toolkit ecosystem.
A Globus account is needed so as to run the workflow to reproduce results reported in this paper.
One can either create one using any email address as ID, or through existing organizational login if participated. 
Both are free of charge.
\paragraph{funcX~\cite{chard2020funcx}} is a distributed Function as a Service (FaaS) platform that enables flexible, scalable, and high performance remote function execution\footnote{\url{https://funcx.readthedocs.io}}. 
It is one of the action providers of the workflow developed in this paper to provision computing resources. 
We installed FuncX server funcX-endpoint version 0.3.2 at all data center systems used in this paper, e.g., via \texttt{pip install funcx\_endpoint}. 
Then we deploy a funcX endpoint by running \texttt{funcx-endpoint configure endpoint-name} and following the authentication instruction.
A UUID will be generated if succeeded, then the endpoint UUID is the unique ID that will be supplied as an argument to the workflow to provision computing resources associated with the endpoint to train the DNN. 

Similarly, we need to install FuncX client (i.e., using \texttt{pip install funcx}) of the same release version on the machine that the user will initialize and start running the workflow.
The client version is needed to register functions to FuncX service so as to supply the function ID to the workflow.

\paragraph{Globus Automate}
The Client SDK of globus Automate provides Python interface for working with Globus Automate, primarily Globus Flows\footnote{\url{https://globus-automate-client.readthedocs.io}}.
We need to install the SDK on the machine that will be used to built and/or run the workflow via \texttt{pip install globus-automate-client}.

\subsection{Run the Proposed Workflow} The implementation of the proposed workflow is open source at \url{https://github.com/AISDC/DNNTrainerFlow} as a Jupyter notebook. \texttt{DNN-Train-Flow.ipynb}. 
There are three major components to build the workflow detailed as follows.
\begin{itemize}
\item Flow definition which basically defines the order and data dependencies (i.e., arguments of each action) of each action. This can be defined using a Python dictionary and the sample can be found from the open source repository. 
\item A self-contained (all modules should be imported within the function) Python function registered to FuncX service to train the model. 
\item Arguments of the workflow which basically contains the full path to the train data and the full path to receive the trained DNN. All arguments are organized in a Python dictionary.   
\end{itemize}
Once the workflow is built and registered to the Globus Flow service, only the returned ID is needed to run it as many times as needed with augments supplied, i.e., similar as running a function with different arguments. 
So we build the workflow once and then share it with user to run it with their own arguments to (re)train their model with supplied data and receive the trained model at the specified destination.

\subsection{Reproduce local GPU results}
The BraggNN~\cite{liu2020braggnn} model as well as the training dataset that we used to demonstrate the workflow is publicly available at \url{https://github.com/lzhengchun/BraggNN}.
One needs to install pyTorch=1.9.0, h5py=2.10.0 and numpy=1.19.2 to reproduce the local single GPU results reported in \autoref{tbl:time-breakdown} by running the \texttt{main.py} or \texttt{horovodrun -np 8 python main-hvd.py} with default arguments.

Similarly, the code and a sample dataset of CookieNetAE can be downloaded from \url{https://github.com/AISDC/CookieNetAE} and run \texttt{horovodrun -np 1 python main-hvd.py} locally to reproduce training using local resource.

\subsection{Reproduce remote GPU cluster results}
As GPU cluster is the most accessible resource among all three AI systems we used in this paper, here we will detail the steps to reproduce the results reported in this paper about using GPU cluster.
The funcX functions are executed under the same Python environment as the funcx-endpoint installation with default configuration. 
As DNN training, especially the data parallelism or AI system based training, usually needs special Python environment. 
We invoke the training script via system call wrapped in the funcX function.

We used Horovod~\cite{sergeev2018horovod} (v0.22.1, installation guide\footnote{\url{https://horovod.readthedocs.io}}) to make use of multiple GPUs via data parallelism. 
One can then run the workflow by supplying \texttt{horovodrun -np 8 python main-hvd.py} as the funcX function argument for the system call, together with other workflow augments to reproduce results reported in \autoref{tbl:time-breakdown}.
Specific instructions for BraggNN and CookieNetAE are detailed as follows:

\subsubsection{BraggNN}
The repository at \url{https://github.com/lzhengchun/BraggNN} hosts the code and dataset, \texttt{main-hvd.py}, for training BraggNN using data parallelism via Horovod.
The distributed data parallel version of BraggNN using Horovod can support multiple GPUs on multiple nodes.
As we discussed in \S\ref{sec:res-discussion}, one will not see much acceleration when training BraggNN using multiple GPUs.
We leave further optimization effort for future work.

\subsubsection{CookieNetAE}
The Horovod based implementation of CookieNetAE and sample dataset are available at \url{https://github.com/AISDC/CookieNetAE}.
Similar as it for BraggNN, one can reproduce results in \S\ref{sec:res-discussion} by supplying \texttt{horovodrun -np 8 python main-hvd.py} to the workflow argument dictionary.

\subsection{SambaNova and Cerebras}
Non-disclosure agreements with SambaNova and Cerebras prevent us from sharing our code for these two systems at this moment. 
As these AI system also have their own Python environment, the way of integration these AI system with the workflow is exactly the same as it for multi-GPU (i.e., via a system call). 
So, we assume one can reproduce similar results as we reported using any other DNNs if SambaNova and/or Cerebras resource is available.

\section*{Acknowledgements}
This material was based upon work supported by the U.S. Department of Energy, Office of Science, under contract DE-AC02-06CH11357 and Office of Basic Energy Sciences under Award Number FWP-35896.
This research used resources of the Argonne Leadership Computing Facility, a DOE Office of Science User Facility supported under Contract DE-AC02-06CH11357.
The demonstration models were developed with support also for the specific detector development for attosecond methods at LCLS-II under Office of Basic Energy Sciences under Award Number FWP-100498.
Much of the computational resources at SLAC were funded under an LDRD award for EdgeML.
CookieNetAE development was funded under Office of Basic Energy Science award FWP-100498 and FWP-100643.
The authors thank Ryan Chard (Argonne National Laboratory) for assistance with funcX.
We thank the engineering support from the Cerebras Systems Inc. and SambaNova Systems.
We thank three anonymous reviewers for their constructive comments. 

\bibliographystyle{IEEEtran}
\bibliography{refs} 

\section*{Government License}
The submitted manuscript has been created by UChicago Argonne, LLC, Operator of Argonne National Laboratory (``Argonne''). Argonne, a U.S.\ Department of Energy Office of Science laboratory, is operated under Contract No.\ DE-AC02-06CH11357. The U.S.\ Government retains for itself, and others acting on its behalf, a paid-up nonexclusive, irrevocable worldwide license in said article to reproduce, prepare derivative works, distribute copies to the public, and perform publicly and display publicly, by or on behalf of the Government.  The Department of Energy will provide public access to these results of federally sponsored research in accordance with the DOE Public Access Plan. http://energy.gov/downloads/doe-public-access-plan.

\end{document}